\documentclass{llncs}

\usepackage[T1]{fontenc}
\usepackage{graphicx} 
\usepackage{amssymb,amsmath}
\usepackage{multirow}
\usepackage{theorem}
\usepackage{todonotes}
\usepackage{comment}
\usepackage{caption}
\usepackage{subcaption}

\newcommand{\chebionehundret}{{\chebi}$_{100}$}
\newcommand{\chebi}{ChEBI}
\newcommand{\disbin}[2]{#1 \sqcap #2 \equiv \bot}
\newcommand{\tee}{T}

\begin{document}

\title{A fuzzy loss for ontology classification}
%
%
\author{Simon Fl\"ugel\inst{1}\orcidID{0000-0003-3754-9016} \and
Martin Glauer\inst{1}\orcidID{0000-0001-6772-1943} \and
Till Mossakowski\inst{1}\orcidID{0000-0002-8938-5204} \and
Fabian Neuhaus\inst{1}\orcidID{0000-0002-1058-3102}
}
\authorrunning{S. Fl\"ugel et al.}
%
\institute{Otto von Guericke University Magdeburg, Germany \\
\email{\{sfluegel, martin.glauer, till.mossakowski, fneuhaus\}@ovgu.de}
}
\maketitle              

\begin{abstract}
Deep learning models are often unaware of the inherent constraints of the task they are applied to. 
However, many downstream tasks require logical consistency. 
For ontology classification tasks, such constraints include subsumption and disjointness relations between classes.

In order to increase the consistency of deep learning models, we propose a fuzzy loss that combines label-based loss with terms penalising subsumption- or disjointness-violations.
Our evaluation on the {\chebi} ontology shows that the fuzzy loss is able to decrease the number of consistency violations by several orders of magnitude without decreasing the classification performance.
In addition, we use the fuzzy loss for unsupervised learning. 
We show that this can further improve consistency on data from a distribution outside the scope of the supervised training.

\end{abstract}




\section{Introduction}
\label{sec:intro}


Deep learning models have been successfully applied to a wide range of classification tasks over the past years, often replacing hand-crafted features with end-to-end feature learning \cite{bojarski2016end,collobert2011natural}. This approach is based on the assumption that all the knowledge required to solve a specific classification task is available in the data used. These systems are often built for a specific use case. In the case of a classification problem, emphasis is placed on the correct classification of the input data and the success of a system is measured in its ability to correctly perform this task. However, this approach disregards that there are often domain-specific logical constraints between different classification targets.

These logical constraints can be of great importance, as applications, especially those leading to further development, are often based on the assumption that inputs are logically consistent. Imagine a system consisting of two components in an autonomous vehicle. The first component recognises and labels objects using a deep learning model. Based on this output, a rule-based system built by experts determines the direction of travel. A contradictory classification of the first system, e.g. a traffic light as both red and green, or a road user as a pedestrian and as a car, can have fatal consequences, as the control system may not cover such a scenario.

It is therefore important to prime systems towards logical consistency. Learning concepts by example is not optimally suited to adhere to domain-specific constraints out of the box. 
Instead, reliance is placed on the fact that the corresponding constraints are represented in the data and that the model can approximate them accordingly during training. 
However, this approach has significant disadvantages. Firstly, it assumes that there is a sufficiently large amount of data so that the corresponding constraints are well represented. Secondly, the system is deprived of important information that is readily available in the domain. Thirdly, it creates an additional, implicit learning task that is not adequately represented by the loss function.

For many research domains, ontologies exist that define important concepts and their relations via logical constraints \cite{bayerlein2024pmd,booshehri2021introducing,hastings2016chebi}. Ontologies therefore provide a necessary logical axiomatisation that can be used to check the consistency of models and to prime them for consistency.
For instance, the subsumption relation \textit{A is-a B} requires that every entity classified as \textit{A} is also classified as \textit{B}. 
Usually, this knowledge is not explicitly given to machine learning models trained on concepts from an ontology. 
Instead, it can only be derived implicitly from seeing a large enough number of \textit{A} samples that are also \textit{B} samples.

The aim of this paper is to integrate symbolic knowledge from ontologies into the learning process of a machine learning model. 
To this end, in Section~\ref{sec:sem-loss}, we present a \textit{fuzzy loss} that extends regular loss functions by additional terms that ensure the model's coherence with ontological constraints. 
In Section~\ref{sec:exp-setup}, we introduce a classification task on the {\chebi} ontology and appropriate evaluation metrics. 
These are used in Section~\ref{sec:results}, where we evaluate different fuzzy loss variants.
The results are discussed in Section~\ref{sec:discussion} and a conclusion is drawn in Section~\ref{sec:conclusion}.

\section{Related Work}~\label{sec:related-work}
A well-studied field within Machine Learning are hierarchical multi-label classification tasks, in which labels are structured in a hierarchy, similar to subsumption relations in an ontology. 
However, it is usually assumed that each class only has one superclass, which allows the assigning of hierarchy levels. 
Many models use these levels directly in their architecture~\cite{cerri2014hierarchical,wehrmann2018hierarchical}.
In ontologies such as {\chebi}, many classes have multiple superclasses, which makes the assignment of hierarchy levels non-trivial.
In addition, ontologies include different kinds of logical relations between classes, such as disjointness or parthood relations.
Therefore, our task requires a more general approach towards ensuring logical consistency.

Neural networks are, in particular during training, prone to making mistakes that may result in logically inconsistent predictions. 
An image recognition system may, for example, classify the same picture as a cat and a dog, although the classes ``cat'' and ``dog'' are known to be disjoint.
Neuro-symbolic systems \cite{DBLP:series/faia/369} integrate deep learning systems which such background knowledge, pushing the classification towards results that are consistent with this knowledge.
%
This allows the priming of a learning system with prior knowledge. 

The training process of neural networks is usually based on a form of gradient descent. 
Consequently, in order to allow answers as truth values $\{0,1\}$, one must allow arbitrary predictions from $[0,1]$ in order to remain differentiable. 
This naturally leads to an interpretation of these values as values from a many-valued logic such as fuzzy logic or probabilistic logic.

The probabilistic interpretation has been used e.g.\ in
DeepProbLog~\cite{manhaeve2018deepproblog}, and in the semantic loss
of Xu et al.~\cite{xu2018semantic}. Semantic loss in based on weighted
model counting (WMC) \cite{DBLP:journals/ai/ChaviraD08}, a technique
to efficiently support Bayesian inference. Xu et al.\ apply this
technique to neural networks. They interpret outputs of a neural
network (once normalised into [0,1]) as probabilities of Boolean
variables. The semantic loss integrates propositional constraints over
these Boolean variables into the loss function. However, Xu et al.\ make the simplifying assumption that these variables
are all independent of each other, which means that neither a more
complex Bayesian inference, nor a more complex encoding of random
variables into logical propositions that also cover conditional
probabilities (as done in
\cite{DBLP:journals/ai/ChaviraD08}) are necessary.

We do not follow this probabilistic approach here, mainly because in
our case, neural outputs are better interpreted as confidence values
than as probabilities. Also, the independence assumption is not
realistic --- after all, we want to use background knowledge to
express certain dependencies. Hence, we employ fuzzy logic.
Indeed, there have been many approaches that aim to combine fuzzy systems and neural networks \cite{kruse1998neuro,zhang2004extended}. 
These systems are particularly useful when training data is limited. 
In a recent work~\cite{glauer2022esc,hastings2023predicting}, we applied an ontology-based neuro-fuzzy controller. 
The approach in this paper is inspired by this work, in which we also apply a semantic penalty system to ensure logically sound rules.

Most prominently, a form of fuzzy loss has been employed by
Logic Tensor Networks
(LTNs, \cite{DBLP:journals/ai/BadreddineGSS22}). They use such a loss
to train neural predicates to maximise satisfiability of a background
theory.  Also Logical Neural Networks (LNNs, \cite{riegel2020logical})
employ fuzzy logic.  Using upper and lower bounds instead of exact truth
values, they model uncertainty. Their fuzzy loss measures a
fuzzy form of inconsistency, pushing the network to avoid logical
inconsistencies between the classification and the background theory.

\cite{DBLP:journals/ai/KriekenAH22} consider different fuzzy operators
and examine their suitability for a fuzzy loss. Here,
we mainly follow their findings, especially the good suitability of
fuzzy logic based on the product t-norm. That said, we also exploit
the flexibility of a fuzzy approach, especially when considering
variants of fuzzy implication below.

Given a class subsumption $A \sqsubseteq B$ in the ontology,
it can be violated by classifying an entity as $A$ but not $B$.
Consequently, using the product t-norm, our loss for such a subsumption
for a given prediction $\hat{y}$ (for an input instance $x\in X$) is defined as
\begin{equation}
  L_{prod}(A \sqsubseteq B, \hat{y}) = \hat{y}_A \cdot (1 - \hat{y}_B)
\label{eq:fuzzy-loss}  
\end{equation}
where for any class $A\in O$ of the ontology $O$,
$\hat{y}_A$ is the component of $\hat{y}$ predicting membership
of the input instance in class $A$.

Interestingly, a fuzzy loss based on the product t-norm
bears a certain similarity to Xu et al.'s semantic loss.
The reason is that fuzzy conjunction is a product in this case,
and probabilistic conjunction is product as well, at least for
independent variables.
In more detail, Xu et al.~\cite{xu2018semantic} 
propose a more general definition of a semantic loss for arbitrary propositional sentences over propositions $V_1,\cdots,V_n$. Given a prediction $\hat{y}$ and a sentence $\varphi$, Xu et al. define their loss using a variant of weighted model counting. Herein, a model $v$ is weighted by its probability w.r.t. $\hat{y}$, meaning that for each proposition $V_i$, the prediction of the neural network $\hat{y}_i$ is construed to be a probability. This loss can therefore be expressed as 
\begin{equation}
    L'_{Xu}(\varphi, \hat{y}) \propto - \log \sum \limits_{v \models \varphi}\ \prod \limits_{i\in\{1,\ldots,n\}, v \models V_i} \hat{y}_i \prod \limits_{i\in\{1,\ldots,n\}, v \models \neg V_i} (1-\hat{y}_i).
\end{equation}

We can apply the loss defined by Xu et al.\ to a subsumption relation $A \sqsubseteq B$ by interpreting it as an implication $V_A \rightarrow V_B$, where
for any class $C\in O$, $V_C$ is a proposition expressing membership in class $C$. Moreover, we construe (crisp) ground truth label vectors $y$ as propositional models $y:\{V_C\mid C\in O\}\to\{0,1\}$.
Because the satisfaction relation $y\models V_C$ is defined under propositional semantics, $y\models A \rightarrow B$ holds iff $y\models \neg V_A$ or $y\models V_B$ for the corresponding propositions $V_A$ and $V_B$. 
Therefore, given a prediction vector $\hat{y}$,
 the loss is calculated as  $L_{Xu}(A \sqsubseteq B, \hat{y}) :=  L'_{Xu}(V_A \rightarrow V_B, \hat{y})$ and
\begin{equation}
  \begin{array}{rl} L'_{Xu}(V_A \rightarrow V_B, \hat{y}) &\propto - \log \sum \limits_{y\models V_A \rightarrow V_B} \prod \limits_{C\in O, y\models V_C} \hat{y}_C \prod \limits_{C\in O, y\models \neg V_C} (1-\hat{y}_C) \\
\multicolumn{1}{r}{(*)}  &=   - \log \sum \limits_{y\models V_A \rightarrow V_B} \prod \limits_{C\in \{A,B\}, y\models V_C} \hat{y}_C \prod \limits_{C\in \{A,B\},y\models \neg V_C} (1-\hat{y}_C) \\
\multicolumn{1}{r}{(**)} &= - \log ( (1-\hat{y}_A) \cdot (1-\hat{y}_B) + (1-\hat{y}_A) \cdot \hat{y}_B + \hat{y}_A \cdot \hat{y}_B )  \\
    &= -\log ( \hat{y}_A \cdot \hat{y}_B - \hat{y}_A + 1 ) \\
    &= -\log ( 1- \hat{y}_A \cdot (1 - \hat{y}_B))
\end{array}
\label{eq:sem-loss-A-B}
\end{equation}
Here, (*) holds because we only need to consider variables occurring
in the implication $V_A \rightarrow V_B$: for any other variable $V_C$,
make a case split --- 
both cases $V_C=0$ and $V_C=1$ are included, and $\hat{y}_C + (1-\hat{y}_C) =1$. Moreover,  (**) just sums up
the three probabilities of different
possibilities for Boolean variables $V_A$ and $V_B$ to satisfy $V_A\to V_B$.

Compared to Eq.~\ref{eq:fuzzy-loss}, Eq.~\ref{eq:sem-loss-A-B} negates the term
(using $N(t)=1-t$) and applies a negative logarithm. Xu et al.\ introduce
the logarithm to achieve a closer correspondence to cross-entropy loss
functions, while we use the result of the logical evaluation directly.
For comparison, we include the semantic loss defined by Xu et al. in
our evaluation in Section~\ref{sec:results}.

\section{Fuzzy Loss}~\label{sec:sem-loss}

Predictions made by a neural network may contradict a logical theory that underlies the predicted labels. In this work, we aim to incentivise a model to produce logically more consistent predictions by adding an additional term to its loss function. While many types of ontology axioms exist, here, we focus on two types that are widely used and domain independent: subsumption relations, i.e., $A \sqsubseteq B$, and disjointness, i.e., $\disbin{C}{D}$. 
While these axioms are usually interpreted in binary semantics, we need differentiable terms that can be used for training a neural network. We achieve differentiability by applying a fuzzy-logic interpretation~\cite{hajek2013fuzzylogic} to our output values. 
Let $h_C: X \rightarrow [0, 1]$ be a fuzzy membership function for a given class $C$ (leading to predictions $\hat{y}_C=h_C(x)$) and ${\tee}$ a fuzzy t-norm. Our fuzzy loss term for implications is then defined as Reichenbach implication \cite{DBLP:journals/ai/KriekenAH22}:
\begin{equation}~\label{eq:impl-loss}
\begin{split}  L_{{\tee}}(A \sqsubseteq B, x)& := \hat{h}(\neg(A \rightarrow B), x) = \hat{h}(\neg(\neg A \vee B), x) = \hat{h}(A \wedge \neg B, x)\\
 & = {\tee}(h_A(x), 1-h_B(x)).
\end{split}
\end{equation}
This assumes that the fuzzy negation used is a strong negation $N(t) = 1-t$.
\cite{DBLP:journals/ai/KriekenAH22} also discuss other fuzzy implications, but their finding is that the product t-norm for conjunction together with Reichenbach implication are among the best-working fuzzy connectives in a machine learning context. (Another good candidate is {\L}ukasiewicz implication that we consider below.)

Accordingly, the fuzzy loss term for disjointness is defined as
\begin{equation}~\label{eq:disj-loss}
    L_{{\tee}}(\disbin{C}{D}, x) := \hat{h}(\neg \neg(C \wedge D), x) = \hat{h}(C \wedge D, x) = {\tee}(h_C(x), h_D(x)).
\end{equation}
Intuitively, the fuzzy loss can be interpreted as the degree to which a given prediction violates an ontological constraint.

In Section~\ref{sec:results}, we evaluate loss functions derived from two commonly used t-norms, the product t-norm ${\tee}_{prod}(a,b) = a \cdot b$ and the {\L}ukasiewicz t-norm ${\tee}_{luka}(a,b) = \max(a + b - 1, 0)$.

Let ${x}\in X$ be a sample vector, ${y}$ the corresponding label vector and ${\hat{y}} = [h_C(x)]_{C\in O}$ the vector of predicted labels.
Based on the loss terms given in Eqs~\ref{eq:impl-loss} and~\ref{eq:disj-loss}, we define our loss function as follows:
\begin{equation}~\label{eq:sem-loss}
    \begin{array}{lcl}
    L_{{\tee}}({x}, {y}) = & & \displaystyle{L_{base} ({y}, {\hat{y}})} \\ 
    &+& \displaystyle{{w_{impl}}   \sum \limits_{A \sqsubseteq B} {\tee}(\hat{y}_A, 1-\hat{y}_B)} \\ 
    &+& \displaystyle{{w_{disj}}  \displaystyle\sum \limits_{\disbin{C}{D}} {\tee}(\hat{y}_C, \hat{y}_D)}
    \end{array}
\end{equation}
The $L_{base}$ term refers to the supervised loss used to train the model on the classification task. 
The weights $w_{impl}$ and $w_{disj}$ are intended to adjust the importance of the fuzzy loss terms in relation to the base loss and to compensate for the different prevalences of the axiom types in the ontology. 
In general, we expect the number of subsumption and disjointness relations in an ontology to vary based on the hierarchy depth and number of disjointness axioms available.
Therefore, these weights have to be adjusted based on the task at hand.

\subsection{Balanced implication loss}~\label{sec:balanced-impl-loss}
The loss terms for implication face an imbalance issue: 
Since the classes on the left-hand side of each implication are subclasses of the right-hand side classes, they necessarily have fewer members in the ontology and therefore fewer labels in a given dataset.
Since we include transitive subsumption relations as well, the difference may be drastic, with some left-hand side classes representing only a small fraction of the right-hand side class.
Therefore, in case of violations, it might be relatively inexpensive for the model to predict non-membership for classes further down in the hierarchy entirely. 
This strategy results in a low number of implication violations, since such classes appear mostly on the left-hand side of implications, and a low supervised loss, due to the lack of positive samples.

However, this behaviour is clearly not in our interest. If class membership
in smaller classes is not predicted correctly, the most important information is lost. After all, membership in classes higher up in the hierarchy is more common and therefore less interesting.
To counter-balance this effect, the \textit{balanced implication loss} has a lower gradient for the left-hand class and a higher gradient for the right-hand class instead of applying the same gradient to both classes.
Practically, this is achieved with two additional parameters $k > 1$ and $\epsilon > 0$: 
\begin{equation}
    L^B_{{\tee}}(A \sqsubseteq B, x; k, \epsilon) = {\tee}\left(\frac{((h_A(x)+\epsilon)^{1/k}-\epsilon^{1/k})}{((1+\epsilon)^{1/k}-\epsilon^{1/k})}, (1-h_B(x))^k\right)
\end{equation}
$\epsilon$ is a small constant that is added to $h_A(x)$ to avoid an infinite gradient at $h_A(x) = 0$. 
The additional $\epsilon$-terms adjust the loss so that $L^B_{{\tee}} = 0$ if $h_A(x) = 0$ and $L^B_{{\tee}} = 1$ if $h_A(x) = 1$ and $h_B(x) = 0$.
In our evaluation, we will use $\epsilon = 0.01$.
The parameter $k$ modifies the loss term such that, in the maximal violation case of $h_A(x) = 1$ and $h_B(x) = 0$, the gradient is larger for $h_B$ than for $h_A$. 
For instance, for the product ${\tee}$-norm, $\frac{\partial L^B_{prod}}{\partial h_A(x)}\big|_1 = \frac{1}{k}$ and $\frac{\partial L^B_{prod}}{\partial h_B(x)}\big|_0 = -k$.

The regular implication loss can be seen as a specialised version of this balanced implication loss where $k = 1$ and $\epsilon = 0$.

\section{Experimental setup}~\label{sec:exp-setup}
We evaluate the fuzzy loss for a classification task in the {\chebi} ontology. 
This task has been studied in our previous work and a deep learning-based approach for the {\chebi} classification task has been developed~\cite{glauer2023interpretable,glauer2023neuro,hastings2021learning}.
In all evaluations, we train an ELECTRA model~\cite{clark2020electra} for a hierarchical multi-label classification task in which {\chebi} classes act as labels and molecules as instances.
For a detailed description of the approach, we refer to~\cite{glauer2024chebifier}. 
Here, we just provide an overview. 
The source code for our implementation is available on GitHub~\footnote{\url{https://github.com/ChEB-AI/python-chebai}}.

\subsection{Datasets}~\label{sec:dataset}
Our setup draws data from two sources.
Labelled data is taken from the {\chebi} ontology~\cite{degtyarenko2008chebi,hastings2016chebi}, while additional unlabelled data is sourced from the PubChem database~\cite{kim2023pubchem}.
All datasets are available on Kaggle~\footnote{\url{https://www.kaggle.com/datasets/sfluegel/chebai-semantic-loss}}.

In all datasets, we use the SMILES (Simplified Molecular Input Line Entry System)~\cite{weininger1988smiles}, a common string representation for chemical structures. It encodes molecules as sequences in which characters represent atoms and bonds, with additional notation for branches, rings and stereoisomerism.

For the labelled data, we use version 231 of {\chebi}, which contains 185 thousand SMILES-annotated classes.
Out of these classes, we form the {\chebionehundret} dataset by attaching all superclasses as labels which have at least 100 SMILES-annotated subclasses.
The transitive closure of subsumption relations between the labels is used for the fuzzy loss.
Disjointness axioms for {\chebi} are provided by an additional ontology module\footnote{\url{https://ftp.ebi.ac.uk/pub/databases/chebi/ontology/chebi-disjoints.owl}}. 
Here as well, we take the subsumption closure of all disjointness relations between label-classes. I.e., for each pair of disjoint classes $\disbin{C}{D}$ and their subclasses $A \sqsubseteq C$ and $B \sqsubseteq D$, we also use $\disbin{A}{B}$ for the loss function.
In total, this provides us with 997 labels, 19,308 implication loss terms and 31,416 disjointness loss terms.

From PubChem, we have sourced two distinct unlabelled datasets. 
The first is used during training while the second one, \textit{PubChem Hazardous}, is only used in the evaluation.
The Hazardous dataset includes SMILES strings for chemicals that are annotated with a class from the Globally Harmonized System of Classification and Labelling of Chemicals (GHS)~\cite{ghsrev10}. 
The GHS covers different kinds of health, physical and environmental hazards and has been developed by the United Nations as a standard for labelling hazardous chemicals and providing related safety instructions.
From this, we have removed all SMILES strings that also appear in the labelled dataset.
In our evaluation, we use this dataset to test model performance for a data distribution outside the learning distribution.

For the unlabelled training dataset, we have randomly selected 1 million SMILES strings from PubChem. 
This set has been split into groups of 10,000. From each group, we have selected the 2,000 SMILES strings with the lowest similarity score, resulting in 200,000 instances.
The similarity score used is the sum of pairwise Tanimoto similarities between the RDKit fingerprints of a given SMILES string and all other SMILES strings in the group. 
This way, we ensure with limited computational expense that our dataset covers a broad range of chemicals.

The overall training set (consisting of labelled and unlabelled data) is used for two tasks: 
Firstly, a pretraining step prior to the training on the {\chebionehundret} dataset which is shared by all models in our evaluation. 
And secondly, a semi-supervised training, in which we train a model simultaneously on labelled and unlabelled data. While labelled data contributes to both prediction loss and fuzzy loss, unlabelled data contributes to the fuzzy loss only, that is, only the consistency of the prediction with the ontology is measured.

In our evaluation, we will compare semi-supervised training to training on only labelled data.

\subsection{Loss function}
In order to apply the fuzzy loss function from Eq.~\ref{eq:sem-loss}, we need to choose a classification loss $L_{base}$ and assign the weights $w_{impl}$ and $w_{disj}$.

For the classification loss, we have chosen a weighted binary cross-entropy loss:
\begin{equation}
    L_{base}({x}, {y}) =  \sum_{C \in O} w_C y_C \cdot \log h_C(x) + (1 - y_C) \cdot \log (1 - h_C(x))
\end{equation}
Here, $w_C$ is a weight assigned to positive entries based on the class $c$. 
These weights are used to increase the importance of classes with fewer members in an imbalanced dataset. 
We apply the scheme introduced by \cite{cui2019class} with $\beta=0.99$ and normalize the weights:
\begin{equation}
    w_C = \frac{w_C' \cdot |O|}{\sum_{C' \in O} w'_{C'}} \text{ where } w_C' = \frac{1 - \beta}{1 - \beta^{|O|}}
\end{equation}

For the fuzzy loss terms, although the number of disjointness terms (31,416) is larger than the number of implication terms (19,308), we have chosen the weights $w_{impl} = 0.01$ and $w_{disj} = 100$.
This is motivated by preliminary experiments in which the implication loss was larger than the disjointness loss by several orders of magnitude.


\subsection{Violation metrics}
In order to quantify the consistency of model predictions with the ontology, we introduce a notion of true positives (TPs) and false negatives (FNs) for consistency violations. 
In this context, all pairs of {\chebionehundret}-labels are considered as violation-labels.
These labels are positive if an explicit subsumption / disjointness relation between both classes exists in {\chebi}. 
Individual predictions are converted into truth values according to a threshold of $0.5$ and the resulting truth values are compared against the label-pairs.

Given a sample $x$, we define the number of TPs as
\begin{equation}
    \#TP_{impl}(x) = | \{(A, B): A \sqsubseteq B \wedge h_A(x) > 0.5 \wedge h_B(x) > 0.5 \}|
\end{equation}
and the number of FNs as
\begin{equation}
    \#FN_{impl}(x) = | \{(A, B): A \sqsubseteq B \wedge h_A(x) > 0.5 \wedge h_B(x) \leq 0.5 \}|.
\end{equation}
For disjointness, the definition is analogous:
\begin{equation}
    \#TP_{disj}(x) = | \{(C, D): \disbin{C}{D} \wedge h_C(x) > 0.5 \wedge h_D(x) \leq 0.5 \}|
\end{equation}
\begin{equation}
    \#FN_{disj}(x) = | \{(C, D): \disbin{C}{D} \wedge h_C(x) > 0.5 \wedge h_D(x) > 0.5 \}|.
\end{equation}

This definition does not take the cases $h_A(x) \leq 0.5$ or $h_C(x) \leq 0.5$ into account which could be considered as true positives as well since they do not contradict the ontology axioms. 
However, since these cases do not require an active prediction, we consider them as "consistent by default" and as less relevant for our evaluation.
Note that, although disjointness axioms are symmetric, this non-symmetric metric requires that we consider both "directions" of the axiom: 
If $\disbin{C}{D}$, $h_C(x) \leq 0.5$ and $h_D(x) > 0.5$, we do not count $(C,D)$ as a true positive, but instead count $(D,C)$.
This also means that $\#FN_{disj}(x)$ will necessarily be even since for every false negative $(C,D)$, there is another false negative $(D,C)$.

Given the numbers of TPs and FNs, we use the false negative rate (FNR), defined as
\begin{equation}
    FNR_{t}(x) = \frac{\#FN_{t}(x)}{\#FN_{t}(x) + \#TP_{t}(x)},
\end{equation}
in our evaluation, $t$ being either $impl$ or $disj$.

\section{Results}~\label{sec:results}

We evaluate four configurations of the fuzzy loss, one using the {\L}ukasiewicz ${\tee}$-norm ${\tee}_{Luka}(x, y) = \max(0, a + b - 1)$ and three using the product ${\tee}$-norm ${\tee}_{prod}(x,y) = x \cdot y$.
For the product ${\tee}$-norm, we include, besides the "standard" variant, one which uses the balanced implication loss $L^B$ described in Section~\ref{sec:balanced-impl-loss} with $k=2$ and the semi-supervised variant trained on a mixed {\chebionehundret} and PubChem dataset (see Section~\ref{sec:dataset}).
For comparison, we also include a configuration using the semantic loss described by Xu et al.~\cite{xu2018semantic} and a baseline configuration trained without fuzzy nor semantic loss.

\begin{table}[tb]
    \centering
    \begin{tabular}{c|c}
        Vocabulary size & 1,400 \\
        Hidden size & 256 \\
        \# attention heads & 8 \\
        \# hidden layers & 6 \\
        \# max. epochs & 200 \\
        learning rate & $10^{-3}$ \\
        Optimizer & Adamax \\
        $w_{impl}$ & 0.01 \\
        $w_{disj}$ & 100 \\
        $\beta$ & 0.99 \\
    \end{tabular}
    \caption{Hyperparameters used during training}
    \label{tbl:hyperparams}
\end{table}
We have conducted a pretraining run on our PubChem dataset and subsequently, 3 fine-tuning runs for each variant. 
In the following, we will only report averages for the 3 runs, the results for individual runs can be found in Appendix~\ref{sec:individual-runs}. 

The hyperparameters shared by all models are given in Table~\ref{tbl:hyperparams}.
We have split the {\chebionehundret} dataset into a training, validation and test set with a 340/9/51 ratio. 
The evaluation has been conducted on the test set using the models with the highest micro-F1 score from each training run.
\begin{table}[tb]
    \centering
    \begin{tabular}{l|cc}
    & {\chebionehundret} & PubChem Hazardous \\  \hline
    baseline & $0.0031 \pm 0.0002$ & $0.0067 \pm 0.005$ \\
    ${\tee}_{Luka}$ & $\mathbf{2.09 \times 10^{-5} \pm 3.3 \times 10^{-6}}$ & $1.49 \times 10^{-5} \pm 2.6 \times 10^{-5}$  \\
    ${\tee}_{prod}$ & $3.18 \times 10^{-5} \pm 1.2 \times 10^{-5}$ & $5.54 \times 10^{-5} \pm 4.6 \times 10^{-5}$ \\
    ${\tee}_{prod}$ (k=2) & $3.74 \times 10^{-5} \pm 1.5 \times 10^{-5}$ & $7.29 \times 10^{-5} \pm 8.8 \times 10^{-5}$  \\
    ${\tee}_{prod}$ (mixed data) & $5.91 \times 10^{-5} \pm 3.2 \times 10^{-5}$ & $\mathbf{1.05 \times 10^{-5} \pm 9.8 \times 10^{-6}}$ \\
    Xu et al. & $3.62 \times 10^{-5} \pm 1.2 \times 10^{-5}$ & $3.93 \times 10^{-5} \pm 2.8 \times 10^{-5}$ \\
    \end{tabular}
    \caption{Average FNR for binary implication violations on the {\chebionehundret} and PubChem Hazardous datasets. The FNR has been calculated separately for each run before averaging. In addition, the table shows the standard deviation between the runs.}
    \label{tbl:fnr-impl}
\end{table}

\begin{figure}[tb]
\centering
\begin{subfigure}{0.49\textwidth}
    \centering
    \includegraphics[width=\textwidth]{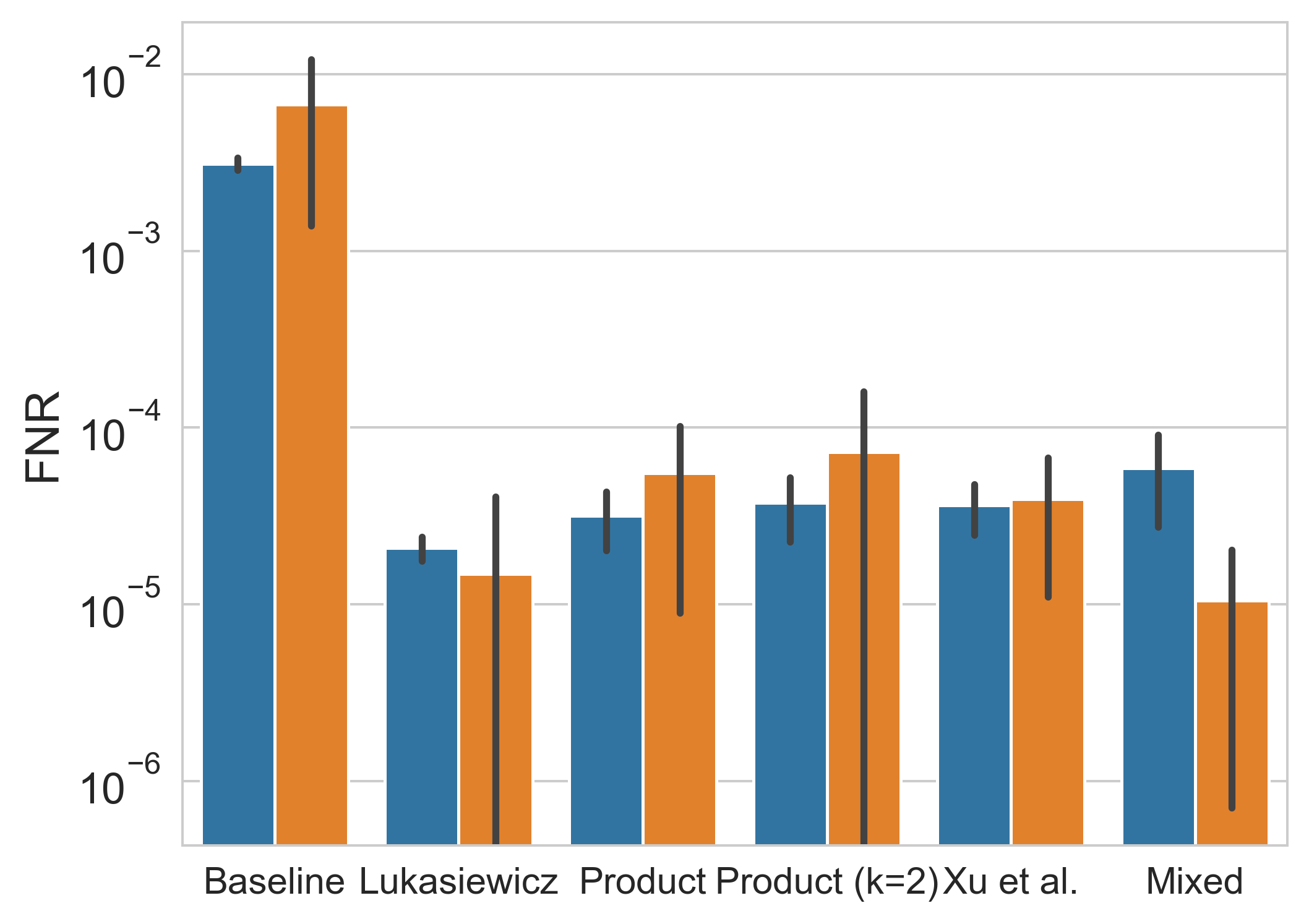}
    \caption{FNR for implication violations on the {\chebionehundret} (left) and PubChem Hazardous datasets. A lower FNR corresponds to more consistent predictions.}
    \label{fig:fnr-impl}
\end{subfigure}
\begin{subfigure}{0.49\textwidth}
    \centering
    \includegraphics[width=\textwidth]{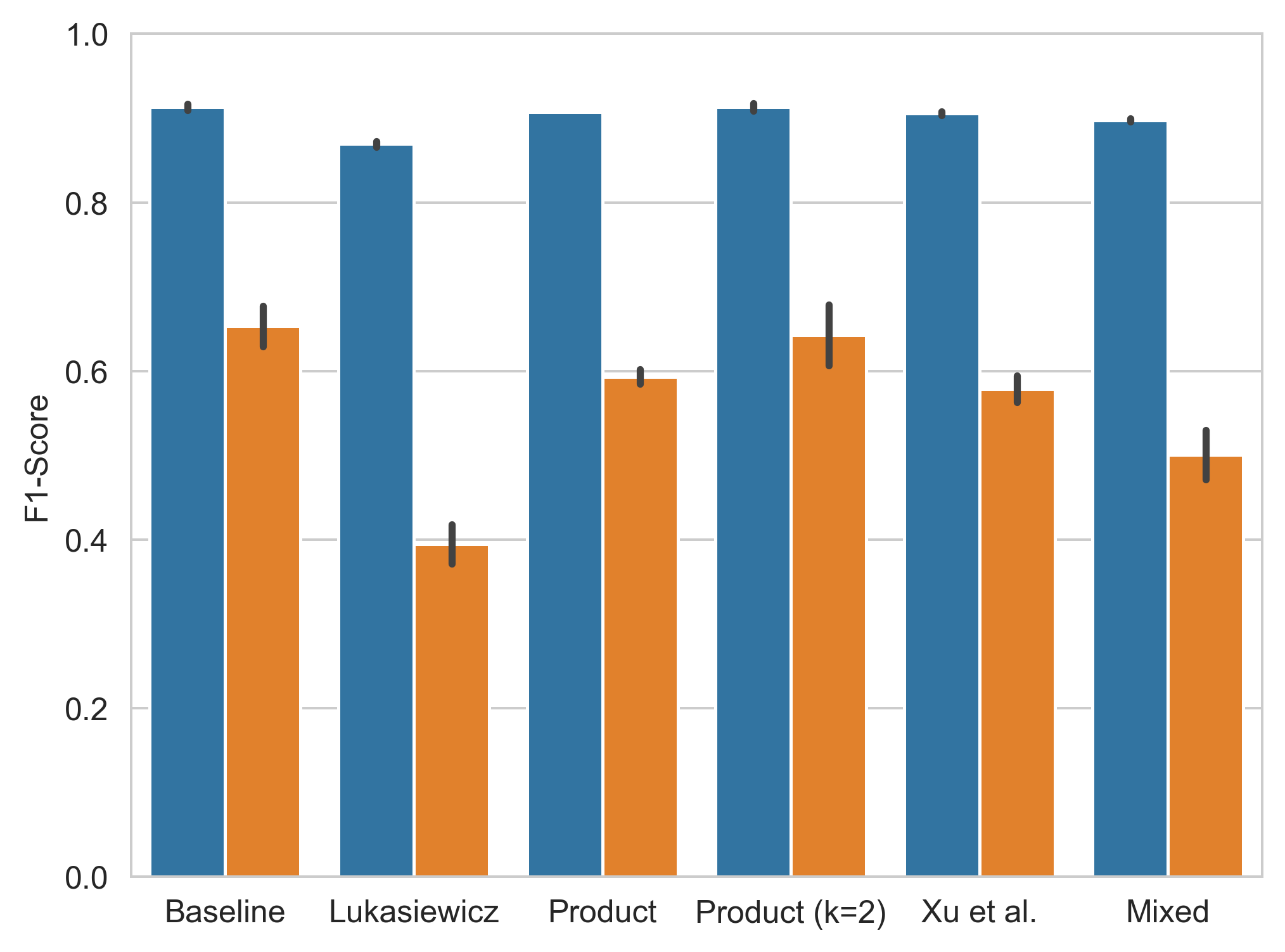}
    \caption{F1-scores on the {\chebionehundret} dataset, aggregated on the micro- (left) and macro-level using a threshold of $t=0.5$. A higher F1-score corresponds to better classification performance.}
    \label{fig:f1-scores}
\end{subfigure}
\caption{Performance of the evaluated models regarding implication violations classification performance. In both figures, the standard deviation is indicated by a black line for each bar.}
\end{figure}

Table~\ref{tbl:fnr-impl} and Figure~\ref{fig:fnr-impl} show the false negative rate (FNR) for implication and disjointness violations on the {\chebionehundret} and the PubChem Hazardous datasets.
It can be seen that all models outperform the baseline by a margin of about two orders of magnitude. 
In absolute terms, this corresponds to $1.3 \times 10^4$ false negatives for the baseline and between $81$ FNs (for ${\tee}_{Luka}$) and $247$ FNs (${\tee}_{prod}$ with mixed data) for the fuzzy loss models. 
The number of true positives is similar for all models (between $3.85 \times 10^6$ and $4.29 \times 10^6$).

For the {\L}ukasiewicz t-norm models, we observe the lowest FNR on {\chebionehundret}.
The models trained with a product t-norm based loss and the semantic loss of Xu et al. have slightly higher FNRs on {\chebionehundret}. 
Regarding the PubChem Hazardous dataset, it is remarkable that, while most models have a smiliar or slightly higher FNR compared to {\chebionehundret}, the models trained with additional PubChem data perform significantly better on PubChem Hazardous.

Regarding the disjointness violations, we have not observed any violations for any model except the baseline models. 
There, we have averages of 171 FNs on {\chebionehundret} and 4 FNs on PubChem Hazardous. 
While these numbers are far below the numbers of TPs ($1.07\times 10^7$ for {\chebionehundret} and $1.12 \times 10^8$ for PubChem Hazardous), they show that the fuzzy loss had a consistency-improving effect. 
For all fuzzy loss variants, the models were able to produce inconsistency-free results regarding disjointness.

\begin{table}[tb]
    \centering
    \begin{tabular}{l|cc|cc}
    & \multicolumn{2}{c}{$t=0.5$} & \multicolumn{2}{c}{$t=t_{max}$} \\
 & Micro-F1 & Macro-F1 & Micro-F1 & Macro-F1 \\
 \hline
    baseline & $\mathbf{0.913} \pm 0.004$ & $\mathbf{0.653} \pm 0.02$ & $\mathbf{0.913} \pm 0.003$ & $\mathbf{0.654} \pm 0.02$ \\
${\tee}_{Luka}$ & $0.869 \pm 0.003$ & $0.395 \pm 0.02$ & $0.870 \pm 0.003$ & $0.405 \pm 0.02$ \\
${\tee}_{prod}$ & $0.907 \pm 0.0003$ & $0.593 \pm 0.009$ & $0.908 \pm 0.0004$ & $0.576 \pm 0.009$ \\
${\tee}_{prod}$ (k=2) & $\mathbf{0.913} \pm 0.004$ & $0.643 \pm 0.04$ & $\mathbf{0.913} \pm 0.005$ & $0.636 \pm 0.04$ \\
${\tee}_{prod}$ (mixed data) & $0.898 \pm 0.002$ & $0.501 \pm 0.03$ & $0.898 \pm 0.002$ & $0.479 \pm 0.02$ \\
Xu et al. & $0.906 \pm 0.002$ & $0.579 \pm 0.02$ & $0.906 \pm 0.002$ & $0.559 \pm 0.02$ \\
 
\end{tabular}
    \caption{F1-scores calculated on labelled data. The micro-F1 aggregates predictions over all classes before calculation the score, while the macro-F1 is the average of the class-wise scores. $t$ refers to the prediction threshold used. $t_{max}$ is the optimal threshold for each model, calculated based on the micro-F1 of the training set for different thresholds. The exact threshold values used are reported in Appendix~\ref{sec:individual-runs}.}
    \label{tbl:f1-scores}
\end{table}

\begin{table}[]
    \centering
    \begin{tabular}{l|cc}
     & Micro-ROC-AUC & Macro-ROC-AUC \\ \hline
         baseline & $0.9969 \pm 0.0003$ & $0.9838 \pm 0.002$ \\
${\tee}_{Luka}$ & $\mathbf{0.9974} \pm 0.0001$ & $\mathbf{0.9842} \pm 0.0007$ \\
${\tee}_{prod}$ & $0.9957 \pm 0.0002$ & $0.9821 \pm 8.5 \times 10^{-5}$ \\
${\tee}_{prod}$ (k=2)  & $0.9936 \pm 0.0008$ & $0.9813 \pm 0.001$ \\
${\tee}_{prod}$ (mixed data)  & $0.9947 \pm 0.0002$ & $0.9754 \pm 0.0007$ \\
Xu et al. & $0.9958 \pm 0.0002$ & $0.9818 \pm 0.0009$ \\
    \end{tabular}
    \caption{ROC-AUC calculated on labelled data. The micro-ROC-AUC aggregates predictions over all classes before calculation the score, while the macro-ROC-AUC is the average of the class-wise scores.}
    \label{tbl:roc-scores}
\end{table}

In addition, we also evaluated the predictive performance of all configurations. 
As can be seen in Table~\ref{tbl:f1-scores} and Figure~\ref{fig:f1-scores}, the F1-score for the models that were trained with fuzzy loss, with exception of the balanced version, is slightly lower than for the baseline models.
This is particularly true for the models that were trained on mixed data or with the {\L}ukasiewicz loss. 

The lower performance of the {\L}ukasiewicz models is linked to an unsuccessful training. 
While the performance of all other models continuously increased during training and converged to the level reported here near the end of allotted 200 epochs, for all 3 {\L}ukasiewicz models, the performance started to drop at approximately 50 epochs into the training. 
Further analysis suggests that the drop during training has been caused by exploding gradients.
Here, we report the results for the best-performing models near the 50 epoch-mark. 
At that point, the performance of the other fuzzy loss models was similar to the {\L}ukasiewicz models.

Table~\ref{tbl:roc-scores} depicts the ROC-AUC metric for all models. All models show very high performance under this metric. Notably, the {\L}ukasiewicz-based model mar\-ginally outperforms all models when considering this metric.

\section{Discussion}~\label{sec:discussion}
Our results indicate that the introduction of a fuzzy loss during training increases the overall logical consistency of predictions significantly. 
However, since the number of consistency violations is relatively low even for the baseline model, one might consider a posteriori processing step that transforms the model output into consistent predictions (e.g., by setting all but the highest output value to 0 for disjointness axioms). 
This can expectedly lead to little or no loss in predictive performance since most predictions are already non-violating and some corrections might even turn wrong predictions into correct ones. 
However, we have only considered the unprocessed predictions in our evaluation. 
We justify this by the intended use cases: 
A model trained on the classification task may further be used in downstream tasks (e.g., prediction of chemical properties~\cite{glauer2023poison}). 
Those require that the model has actually learned the ontology's structure. 
This can only be achieved by giving the model direct feedback during training (as we did with the fuzzy loss) instead of superficially correcting the results.

For most fuzzy loss variants, their increase of consistency comes to the detriment of the actual predictive quality. 
This result seems contradictory at first, as one would assume more consistent results to be better overall. One possible explanation lies in the imbalanced character of ontology-based datasets. The hierarchical relations between labels create a dataset in which a significant imbalance is inevitable and cannot be overcome by sampling procedures in any significant way. A class will always have less members than its parents as long as there is an ontological distinction between them that is also represented in the dataset. Consequently, classes that reside further down in the hierarchy are often significantly smaller than those higher up. For {\chebi} in particular, more specialised classes also require the model to learn relatively complex patterns from a limited amount of samples. 
The corresponding labels receive a relatively small training signal that is then counteracted by the additional loss due to violations. This may render the model unable to learn some smaller classes.

This view is supported by the differences between micro- and macro-F1. 
For all models, the macro-F1 is far lower than the micro-F1. 
This means that many small classes that contribute little to the micro-F1, but receive a stronger weighting in the macro-F1, perform badly. 
For the fuzzy loss variants, with the exception of the balanced fuzzy loss, this gap widens (from $26\%$ for the baseline to $31\%$ for ${\tee}_{prod}$ or $47\%$ for ${\tee}_{luka}$).
This shows that, when the predictive performance decreases, it mostly affects classes with fewer members.

However, a general tendency to make less predictions cannot be observed. 
While all models make a similar amount of predictions on the {\chebionehundret} dataset, the fuzzy loss models make more predictions on average for the PubChem dataset ($9.4$ with product loss, $8.1$ for the baseline).
This shows that the lower F1-score in the final models is not due to a generally more "cautious" behaviour.
Instead, only some classes may get left out while additional (consistent, but wrong) predictions are made for other classes.
Also, the lower performance may be attributed to differences in the learning process, e.g., a less explorative model behaviour or a slower convergence.

\begin{figure}[tb]
    \centering
    \includegraphics[width=\textwidth]{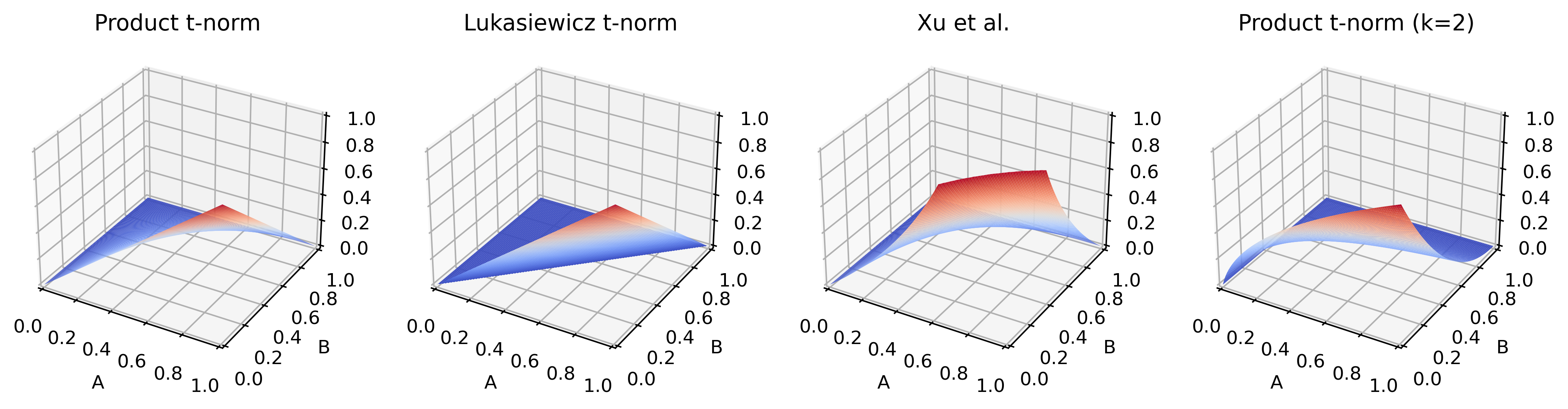}
    \caption{Value of the fuzzy loss variants $L_{prod}$, $L_{luka}$, $L_{Xu}$ and $L^B_{prod}$ with $k=2$ for a subsumption relation $A \sqsubseteq B$ with different values of $h_A(x)$ and $h_B(x)$. $L_{Xu}$ has been cut off at $L_{Xu} = 1$ since $\displaystyle\lim_{p_a\to0, p_b\to1} L_{Xu}(A \sqsubseteq B, p) = \infty$}
    \label{fig:semantic-loss-functions}
\end{figure}

In addition, the results confirm our hypothesis regarding the balanced fuzzy loss. 
Without losing consistency compared to the unbalanced variant, it is able to reach a predictive performance similar to the baseline.
Recall that the main difference between the balanced and unbalanced fuzzy loss is the gradient in cases where for a subsumption relation $A \sqsubseteq B$, the model predicts $h_A(x)$ close to $1$ and $h_B(x)$ close to $0$ (cf. Figure~\ref{fig:semantic-loss-functions}).
There, without the balancing, both classes get the same gradient. 
With balancing, the gradient is stronger for $h_B(x)$ than for $h_A(x)$ (in our experiment, by a factor of approximately 4). 

The balanced fuzzy loss has been successful in pushing the model towards more consistent predictions without pushing it towards predictions that contradict the labels of the classification task.

Notably, the {\L}ukasiewicz-based model marginally outperforms all models when considering the ROC-AUC metric. However, it should be noted that the standard method used to calculate ROC-AUC is based on an analysis of all threshold values on the test data. We see this approach as problematic. The goal of evaluating a machine learning model is to determine whether the model can be used to make accurate predictions. However, we see the threshold as part of this prediction system. The test set should therefore only be used in the final evaluation and no parameters should be chosen to fit a particular test set particularly well. The way in which the ROC-AUC curves are computed dilutes this approach as different thresholds are evaluated using the test data. It is not clear how this analysis can be used to generate a threshold for a good predictive model that is test data agnostic. We conclude, that the threshold should either be fixed for all models or individually determined based on the training set. The performance analysis based on those individual thresholds shows that using those threshold for which each model performed best on the training leads to results that are consistent with those produced by fixed thresholds of $0.5$.

The results also indicate that the inclusion of unlabelled data into the training process does hedge the system against inconsistencies on unseen data. 
This result can be particularly useful in scenarios in which the distribution of features in the dataset is limited. 
Deep learning systems are prone to suffer from out-of-distribution errors, e.g., unpredictable behaviour on data that has not been sampled from the same distribution as the training data. 
Lifting this limitation is often not easy because additional labelling is required. 
The semi-supervised training method presented here can help to alleviate this problem.

\section{Conclusion}~\label{sec:conclusion}
In this work, we have introduced a fuzzy loss function for the task of ontology classification. 
Our fuzzy loss is based on a fuzzy logic interpretation of the ontology subsumption and disjointness relations.
To counteract the loss function's tendency to disincentivise predictions of low-level ontology classes, we have proposed a balanced fuzzy loss variant as well. 
In our evaluation, we have compared different versions of our fuzzy loss (based on either the {\L}ukasiewicz t-norm or the product t-norm) to a baseline model and the semantic loss function proposed by~\cite{xu2018semantic}. 
We have shown that all fuzzy loss variants were able to reduce the number of consistency violations by approximately two orders of magnitude. 

Regarding performance on the classification task, we have seen greater differences between the loss functions. 
Most variants have both a slightly lower micro- and a significantly macro-F1 than the baseline (especially the {\L}ukasiewicz-based variant). 
This indicates that especially the predictive performance of small classes is affected by the fuzzy loss.
Only the balanced fuzzy loss was able to perform on a par with our baseline.

In addition, we have used the fuzzy loss for an additional training task on unlabelled data. 
This allows us to generalise beyond the original data distribution used for supervised training. 
Our evaluation on the Hazardous subset of PubChem shows that this form of training can further improve the consistency of predictions for out-of-distribution data.

Future work will include an improved normalisation to avoid performance issues like we reported for the {\L}ukasiewicz fuzzy loss.
Also, it is possible to extend our approach to other types of ontology axioms, e.g., parthood relations.
Finally, it would be interesting to incorporate our finding of a balanced
implication loss into more general frameworks like LTNs or LNNs.

\section*{Acknowledgements}
This work has been funded by the Deutsche Forschungsgesellschaft (DFG, German Research Foundation) - 522907718.

\appendix

\section{Result for individual runs}~\label{sec:individual-runs}
In Section~\ref{sec:results}, we have presented results as the average and standard deviation out of 3 runs for every configuration. 

\begin{table}
    \centering
    \begin{tabular}{ll|ccc}
 & Dataset & Run1 & Run2 & Run3 \\ \hline
Baseline & {\chebionehundret}  &  $0.0034$ & $0.0031$ & $\mathbf{0.0029}$ \\
 & Hazardous  &  $0.0066$ & $0.012$ & $\mathbf{0.0015}$ \\
${\tee}_{Luka}$ & {\chebionehundret}  &  $2.32 \times 10^{-5}$ & $\mathbf{1.72 \times 10^{-5}}$ & $2.23 \times 10^{-5}$ \\
 &  Hazardous  &  $4.45 \times 10^{-5}$ & $\mathbf{0}$ & $1.79 \times 10^{-7}$ \\
${\tee}_{prod}$ & {\chebionehundret}  &  $\mathbf{1.94 \times 10^{-5}}$ & $3.36 \times 10^{-5}$ & $4.24 \times 10^{-5}$ \\
 & Hazardous  &  $6.26 \times 10^{-5}$ & $\mathbf{5.72 \times 10^{-6}}$ & $9.78 \times 10^{-5}$ \\
${\tee}_{prod}$ (k=2) & {\chebionehundret}  &  $\mathbf{2.41 \times 10^{-5}}$ & $3.46 \times 10^{-5}$ & $5.35 \times 10^{-5}$ \\
 & Hazardous  &  $\mathbf{2.22 \times 10^{-5}}$ & $0.00017$ & $2.25 \times 10^{-5}$ \\
Xu et al. & {\chebionehundret}  &  $4.44 \times 10^{-5}$ & $\mathbf{2.30 \times 10^{-5}}$ & $4.12 \times 10^{-5}$ \\
 & Hazardous  &  $\mathbf{6.68 \times 10^{-6}}$ & $5.58 \times 10^{-5}$ & $5.56 \times 10^{-5}$ \\
${\tee}_{prod}$ (mixed data) & {\chebionehundret}  &  $4.78 \times 10^{-5}$ & $\mathbf{3.46 \times 10^{-5}}$ & $9.48 \times 10^{-5}$ \\
 & Hazardous  &  $\mathbf{2.98 \times 10^{-7}}$ & $1.98 \times 10^{-5}$ & $1.14 \times 10^{-5}$ \\
\end{tabular}
 \caption{FNRs for implication violations on the {\chebionehundret} and PubChem datasets.}
 \label{tbl:fnr-individual-runs}
\end{table}

\begin{table}
    \centering
    \begin{tabular}{lc|ccc}
 & Aggregation & Run1 & Run2 & Run3 \\  \hline
Baseline & micro & $0.909$ & $\mathbf{0.915}$ & $0.915$ \\
 & macro & $0.625$ & $\mathbf{0.669}$ & $0.665$ \\
${\tee}_{Luka}$ & micro & $0.870$ & $0.866$ & $\mathbf{0.873}$ \\
 & macro & $0.399$ & $0.369$ & $\mathbf{0.415}$ \\
${\tee}_{prod}$ & micro & $0.907$ & $\mathbf{0.907}$ & $0.907$ \\
 & macro & $0.584$ & $\mathbf{0.600}$ & $0.597$ \\
${\tee}_{prod}$ (k=2) & micro & $0.915$ & $\mathbf{0.916}$ & $0.908$ \\
 & macro & $0.661$ & $\mathbf{0.666}$ & $0.601$ \\
Xu et al. & micro & $0.904$ & $\mathbf{0.908}$ & $0.905$ \\
 & macro & $0.570$ & $\mathbf{0.597}$ & $0.569$ \\
${\tee}_{prod}$ (mixed data) & micro & $0.895$ & $\mathbf{0.899}$ & $0.899$ \\
 & macro & $0.471$ & $0.502$ & $\mathbf{0.529}$ \\
\end{tabular}
 \caption{F1-scores with micro and macro aggregation, using the threshold $t=0.5$.}
 \label{tbl:f1-individual-runs}
\end{table}

\begin{table}
    \centering
    \begin{tabular}{ll|ccc}
& Value & Run1 & Run2 & Run3 \\ 
\hline
 & $t_{max}$ & $0.5$ & $0.5$ & $0.45$ \\
Baseline & Micro-F1 & $0.909$ & $\mathbf{0.915}$ & $0.915$ \\
 & Macro-F1 & $0.625$ & $\mathbf{0.669}$ & $0.668$ \\
 \hline
 & $t_{max}$ & $0.5$ & $0.45$ & $0.45$ \\
${\tee}_{Luka}$ & Micro-F1 & $0.870$ & $0.866$ & $\mathbf{0.873}$ \\
 & Macro-F1 & $0.399$ & $0.386$ & $\mathbf{0.429}$ \\
 \hline
 & $t_{max}$ & $0.65$ & $0.65$ & $0.65$ \\
${\tee}_{prod}$ & Micro-F1 & $0.908$ & $\mathbf{0.908}$ & $0.908$ \\
 & Macro-F1 & $0.566$ & $\mathbf{0.582}$ & $0.580$ \\
 \hline
 & $t_{max}$ & $0.55$ & $0.6$ & $0.55$ \\
${\tee}_{prod}$ (k=2) & Micro-F1 & $0.915$ & $\mathbf{0.917}$ & $0.908$ \\
 & Macro-F1 & $\mathbf{0.657}$ & $0.656$ & $0.595$ \\
 \hline
 & $t_{max}$ & $0.7$ & $0.65$ & $0.65$ \\
Xu et al. & Micro-F1 & $0.905$ & $\mathbf{0.909}$ & $0.905$ \\
 & Macro-F1 & $0.547$ & $\mathbf{0.580}$ & $0.551$ \\
 \hline
 & $t_{max}$ & $0.6$ & $0.6$ & $0.7$ \\
${\tee}_{prod}$ (mixed data) & Micro-F1 & $0.896$ & $\mathbf{0.899}$ & $0.899$ \\
 & Macro-F1 & $0.457$ & $0.487$ & $\mathbf{0.495}$ \\

\end{tabular}
 \caption{F1-scores with micro and macro aggregation, using the threshold $t_{max}$. $t_{max}$ is the threshold for which the maximum micro-F1 was reached on the training set.}
 \label{tbl:f1-individual-runs-tmax}
\end{table}

Tables~\ref{tbl:fnr-individual-runs}, ~\ref{tbl:f1-individual-runs} and ~\ref{tbl:f1-individual-runs-tmax} show the FNR and and F1-scores for the individual runs.
For the FNR, it can be observed that the results vary significantly between runs, especially on the Hazardous datasets. 
This is likely due to the small scale we are considering:
On the {\chebionehundret}, a FNR of $2.5 \times 10^{-5}$ roughly corresponds to about 100 observed false negatives over the whole test set. I.e., out of 19 thousand samples, each of which had 19 thousand subsumption relations that could have resulted in a false negative, only 100 actually are false negatives.
Therefore, slight changes in the predictive performance can have a significant impact on the false negative rate.

The F1-scores are more stable overall, with a range of less than one percent for the micro aggregation and up to six percent for the macro aggregation.

\bibliographystyle{splncs04}
\bibliography{main}

\end{document}